\documentclass[review]{elsarticle}

\makeatletter
\def\ps@pprintTitle{%
 \let\@oddhead\@empty
 \let\@evenhead\@empty
 \def\@oddfoot{}%
 \let\@evenfoot\@oddfoot}
\makeatother

\usepackage{lineno,hyperref}
\modulolinenumbers[5]

\journal{Journal of Information Processing and Management}

\biboptions{authoryear}

\usepackage{booktabs}

\usepackage{amsmath}
\usepackage{amssymb}
\usepackage{bm}

\usepackage{dashbox}

\usepackage{enumitem}

\usepackage{graphicx}

\usepackage{rotating}

\usepackage{color}

\begin{document}

\begin{frontmatter}

\title{TDAM: a Topic-Dependent Attention Model \\for Sentiment Analysis}

\author{Gabriele Pergola\corref{correspondingauthor}}
\ead{gabriele.pergola@warwick.ac.uk}
\author{Lin Gui\corref{}}
\author{Yulan He\corref{correspondingauthor}}
\ead{y.he@cantab.net}
\cortext[correspondingauthor]{Corresponding authors.}

\address{University of Warwick, Coventry, UK}

\begin{abstract}
We propose a topic-dependent attention model for sentiment classification and topic extraction. Our model assumes that a global topic embedding is shared across documents and employs an attention mechanism to derive local topic embedding for words and sentences. These are subsequently incorporated in a modified Gated Recurrent Unit (GRU) for sentiment classification and extraction of topics bearing different sentiment polarities. Those topics emerge from the words' local topic embeddings learned by the internal attention of the GRU cells in the context of a multi-task learning framework. In this paper, we present the hierarchical architecture, the new GRU unit and the experiments conducted on users' reviews which demonstrate classification performance on a par with the state-of-the-art methodologies for sentiment classification and topic coherence outperforming the current approaches for supervised topic extraction. In addition, our model is able to extract coherent aspect-sentiment clusters despite using no aspect-level annotations for training.
\end{abstract}


\end{frontmatter}


\section{Introduction}

In recent years, attention mechanisms in neural networks have been widely used in various tasks in Natural Language Processing (NLP), including machine translation \citep{Bahdanau2014,luong2015effective,vaswani2017attention}, image captioning \citep{xu2015show}, text classification \citep{Yang16,Chen16,ma2017interactive} and reading comprehension \citep{hermann2015teaching,wang2017gated}. Attention mechanisms are commonly used in models for processing sequence data that instead of encoding the full input sequence into a fixed-length vector learn to ``attend" to different parts of the input sequence, based on the task at hand. This is equivalent to giving the model the access to its internal memory which consists of the hidden states of the sequence encoder. Typically soft attention is used which allows the model to retrieve a weighted combination of all memory locations. 

\begin{figure}[!t]
\centering
\includegraphics[width=0.65\textwidth]{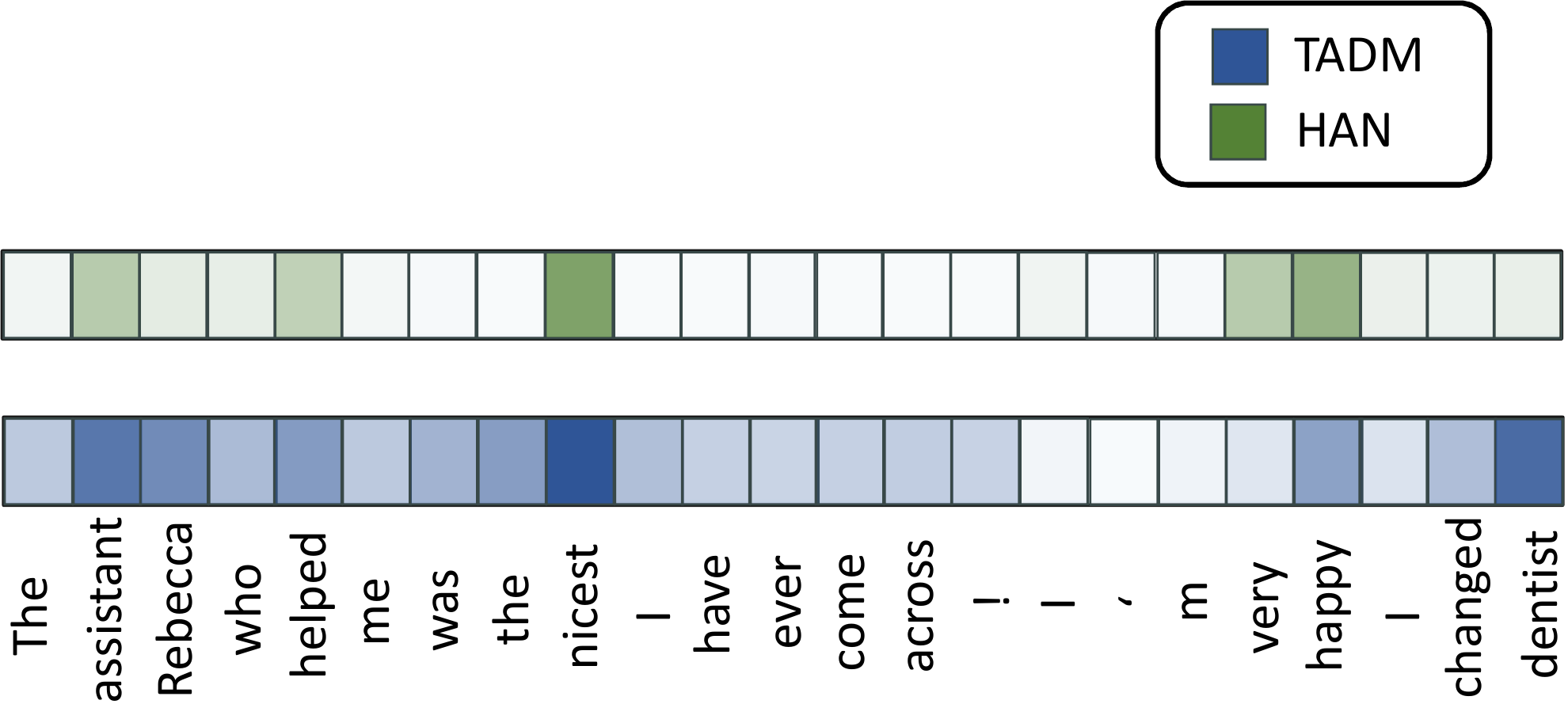}
\caption{Attention weights from the \textit{Topic-Dependent Attention Model} (TDAM) and \textit{Hierarchical Attention Network} (HAN) \citep{Yang16}. TDAM  highlights and gives more relevance to both sentiment and topical words.}

\label{fig:att_weights}
\end{figure}

One advantage of using attention mechanisms is that the learned attention weights can be visualized to enable intuitive understanding of what contributes the most to the model's decision. For example, in sentiment classification, the visualization of word-level attention weights can often give us a clue as to why a given sentence is classified as positive or negative. Words with higher attention weights can be sometimes indicative of the overall sentence-level polarity (for example, see Figure \ref{fig:att_weights}). This inspires us the development of a model for the extraction of polarity-bearing topics based on the attention weights learned by a model.

However, simply using the attention weights learned by the traditional attention networks such as the Hierarchical Attention Network (HAN) \citep{Yang16} would not give good results for the extraction of polarity-bearing topics, since in these models the attention weight of each word is calculated as the similarity between the word's hidden state representation with a context vector shared across all the documents. There is no mechanism to separate words into multiple clusters representing polarity-bearing topics.

Therefore, in this paper, we propose a novel Topic-Dependent Attention Model (TDAM)\footnote{\url{https://github.com/gabrer/topic_dependent_attention_model}} in which a global topic embedding (i.e., a matrix with $K$ topic vectors) is shared across all the documents in a corpus and captures the global semantics in multiple topical dimensions. When processing each word in an input sequence, we can calculate the similarity of the hidden state of the word with each topic vector to get the attention weight along a certain topical dimension. By doing so, we can subsequently derive the local topical embedding for the word by the weighted combination of the global topic embeddings, indicating the varying strength of the association of the word with different topical dimensions. We use Bidirectional Gated Recurrent Unit (BiGRU) to model the input word sequence; we modify the GRU cells to derive a hidden state for the current word which simultaneously takes into account the current input word, the previous hidden state and local topic embedding.

Our proposed formulation of topical attention is somewhat related to the consciousness prior proposed in \citet{Bengio17-CPrior} in which the conscious state value corresponds to the content of a thought and can be derived by a form of attention selecting a ``small subset of all the information available" from the hidden states of the model. Analogously, we first assume the corpus is characterized by a global topic embedding. Then, we learn how to infer the local topic mixture for each analyzed word/sentence combining hidden states and global topic embedding with attention.

\begin{figure}[!ht]
\centering
\includegraphics[width=0.85\textwidth]{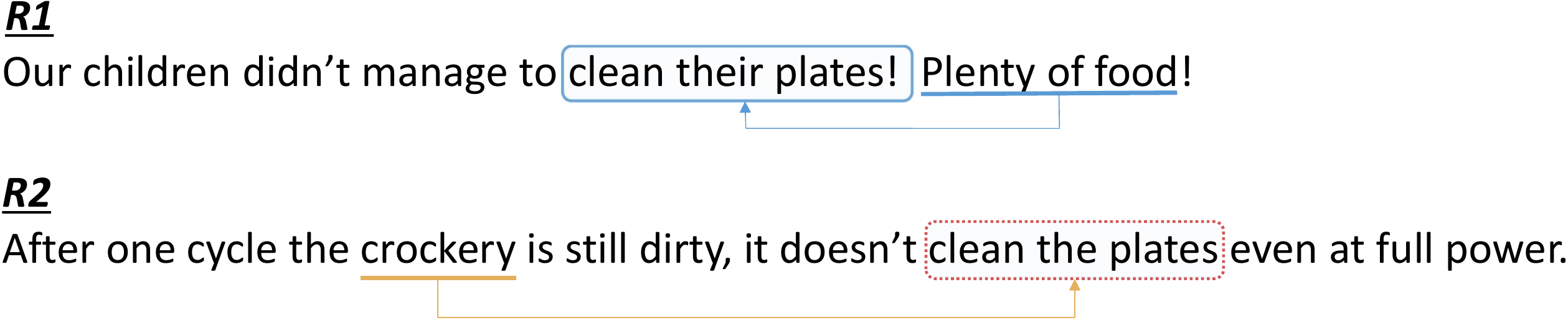}
\caption{An example of topics bearing polarities.}
\label{fig:intro_example}
\end{figure}

In this paper, we describe TDAM and present its application to sentiment classification in reviews by a hierarchical and multi-task learning architecture. The aim is to evaluate a review's polarity by predicting both the rating and the domain category of the review (e.g. \textit{restaurant}, \textit{service}, \textit{health}, etc.). Often these reviews contain statements that can be fully specified only by the contextual topic. To illustrate, in Figure \ref{fig:intro_example} we show two review extracts, one for a restaurant and another for a dishwasher. Interestingly, the same expression ``\emph{not to clean the plates}" can be regarded as positive for food while it bears a negative polarity for kitchen equipment. Thus, it is important to jointly consider both topic and sentiment shared over words for better sentiment analysis.

\noindent In particular, we make the following contributions:
\begin{itemize}
\item We design a neural architecture and a novel neural unit to analyze users' reviews while jointly taking into account topics and sentiments. The hierarchical architecture makes use of a global topic embedding which encodes the shared topics among words and sentences; while the neural unit employs a new internal attention mechanism which leverages the global topic embeddings to derive a local topic representation for words and sentences.

\item We assess the benefit of multi-task learning to induce representations which are based on documents' polarities and domains. Our experiments show that combining the proposed architecture with the modified GRU unit is an effective approach to exploit the polarity and domain supervision for accurate sentiment classification and topic extraction.

\item As a side task to evaluate the sentence representations encoded by TDAM, we extract \textit{aspect-sentiment} clusters using no aspect-level annotations during the training; then, we evaluate the coherence of those clusters. Experiments demonstrate that TDAM achieves state-of-the-art performance in extracting clusters whose sentences share coherent polarities and belong to common domains.
\end{itemize}

To evaluate the performance of our model, we conduct experiments on both Yelp and Amazon review datasets (see \textsection \ref{ss:datasets}). We compare the sentiment classification performance with state-of-the-art models (\textsection \ref{s:acc_results}). Then, visualization of topical attention weights highlights the advantages of the proposed framework (\textsection \ref{ss:ablation_study}). We also evaluate how meaningful are the inferred representations in term of topic coherence (\textsection \ref{ss:topic_evaluation}) and based on their capability to cluster sentences conveying a shared sentiment about a common aspect (\textsection \ref{ss:phrase_clustering}).

\section{Related Work}

\noindent Our work is related to three lines of research.

\textbf{Hierarchical structure for text classification}. Many works have recently proposed to incorporate prior knowledge about the document structure directly into the model architecture to enhance the model's discriminative power in sentiment analysis.  A hierarchical model incorporating user and product information was first proposed by \citet{Tang15} for rating prediction of reviews. 
Similarly, \citet{Chen16} combined user and product information in a hierarchical model using attention \citep{Bahdanau2014}; here, attention is employed to generate hidden representations for both products and users. \citet{Yang16} used a simple and effective two-level hierarchical architecture to generate document representations for text classification; words are combined in sentences and in turn, sentences into documents by two levels of attention.
\citet{Liu18} further empowered the structural bias of neural architectures by embedding a differentiable parsing algorithm. This induces dependency tree structures used as additional discourse information; an attention mechanism incorporates these structural biases into the final document representation.
\citet{Yang2019} introduced Coattention-LSTM for aspect-based sentiment analysis which designs a co-attention encoder alternating and combining the context and target attention vectors of reviews.

\textbf{Combining topics with sequence modeling}. 
There has been research incorporating topical information into the sequence modeling of text or use variational neural inference for supervised topic learning. \citet{Dieng16} developed a language model combining the generative story of Latent Dirichlet Allocation (LDA) \citep{Blei03} with the word representations generated by a recurrent neural network (RNN). 
\citet{Stab18} proposed incorporating topic information into some gates in Contextual-LSTM, improving generalization accuracy on argument mining.
\citet{Abdi19} proposed to directly incorporate word and sentence level features about contextual polarity, type of sentence and sentiment shifts by encoding prior knowledge about part-of-speech (POS) tagging and sentiment lexicons.
 \citet{Kastrati19} enhanced document representations with knowledge from an external ontology and encoded documents by topic modeling approaches.
\citet{Jin18} proposed to perform topic matrix factorization by integrating both LSTM and LDA, where LSTM can improve the quality of the matrix factorization by taking into account the local context of words. 
\citet{Card18} proposed a general neural topic modeling framework which allows incorporating metadata information with a flexible variational inference algorithm. The metadata information can be labels driving the topic inference and used for the classification task, analogous to what proposed in a Bayesian framework by \citet{Blei08} with supervised Latent Dirichlet Allocation (S-LDA).

\textbf{Multi-task learning}. Several variants of multi-task learning with neural networks have been recently used for sentiment analysis.

\noindent\citet{Wu16} proposed a multi-task learning framework for microblog sentiment classification which combines common sentiment knowledge with user-specific preferences. 
\citet{Liu16} employed an external memory to allow different tasks to share information. 
\citet{Liu17} proposed an adversarial approach to induce orthogonal features for each task. \cite{Chen18} applied a different training scheme to the adversarial approach to minimize the distance between feature distributions across different domains. \citet{zhang18} proposed to use an embedded representation of labels to ease the generation of cross-domain features. \citet{Zheng18} proposed to share the same sentence representation for each task which in turn can select the task-specific information from the shared representation using an ad-hoc attention mechanism.
\citet{Wang18} applied multi-task learning for microblog sentiment classification by characterizing users across multiple languages.

\section{Topic-Dependent Attention Model}

\begin{figure*}[t]
\centering
\includegraphics[width=1\textwidth]{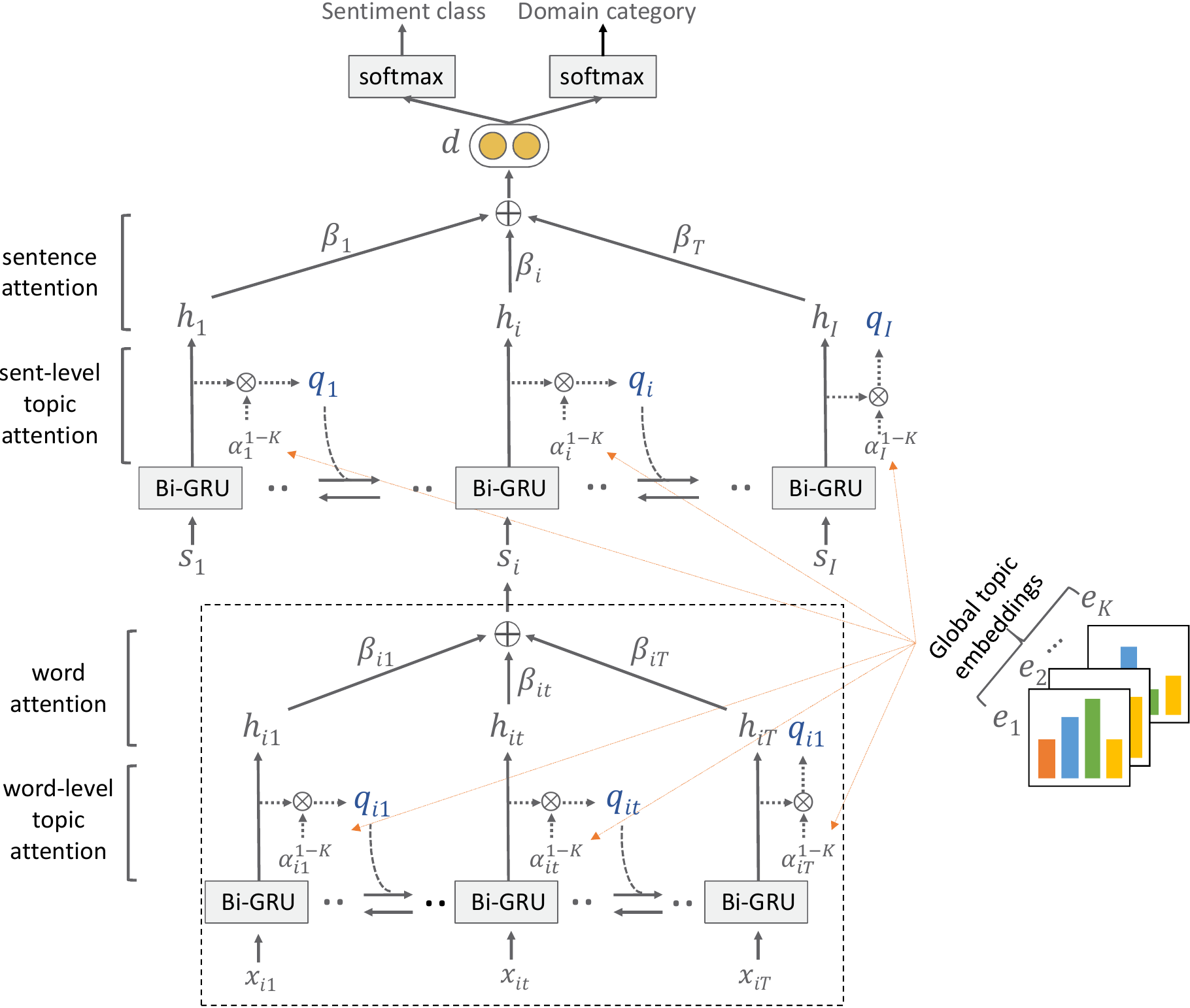}
\caption{Topic-Dependent Attention Model (TDAM).}
\label{fig:TDAM}
\end{figure*}

We illustrate the architecture of our proposed Topic-Dependent Attention Model (TDAM) in Figure \ref{fig:TDAM}, which is a hierarchical and multi-level attention framework trained with multi-task learning. 

Concretely, at the word sequence level (the bottom part of Figure \ref{fig:TDAM}), we add a word-level topic attention layer which computes the local topic embedding of each word based on the global topic embedding and the current hidden state. Such word-level local topic embedding indicates how strongly each word is associated with every topic dimension, which is fed into the Bi-GRU cell in the next time step for the derivation of the hidden state representation of the next word. Bi-GRU is used to capture the topical contextual information in both the forward and backward directions. We then have a word attention layer which decides how to combine the hidden state representations of all the constituent words in order to generate the sentence representation. At the sentence-level, a similar two-level attention mechanism is used to derive the document representation, which is fed into two separate softmax layers for predicting the sentiment class and the domain category. Each of the key components of TDAM is detailed below.

\subsection{Topic-Dependent Word Encoder}
\label{sec:topic_word_encoder}
Given a word sequence $\boldsymbol{x}_i = (x_{i1}, \dots, x_{iT})$, where $x_{it} \in \mathbb{R}^d$ is a word embedding vector with $d$ dimensions, we use Bi-GRU to encode the word sequence.
The hidden state at each word position, $h_{it}$, is represented by the concatenation of both forward and backward hidden states, $h_{it} = [\overrightarrow{h_{it}}, \overleftarrow{h_{it}}]$, which captures the contextual information of the whole sentence centred at $x_{it}$.

We assume there are $K$ global topic embeddings shared across all documents, where each topic has a dense and distributed representation, $e_k \in \mathbb{R}^n,$ with $k=\{1,...,K\}$, which is initialized randomly and will be updated during model learning.

At each word position, we can calculate the word-level topic weight by measuring the distance between the word vector and each global topic vector. We first project $h_{it}$ using a one-layer MLP and then compute the dot products between the projected $h_{it}$ and global topic vectors $e_k, k=\{1,...,K\}$ to generate the weight of local topic embedding for the corresponding word position\footnote{We drop the bias terms in all the equations in our paper for simplicity.}:
\begin{alignat}{2}
u_{it}  &= \mbox{tanh}(W_w h_{it}) \\
\alpha_{it}^k &= \mbox{softmax}(u_{it}^\intercal e_{k})
\end{alignat}

\noindent where $W_w \in \mathbb{R}^{n\times n}$ and $k \in \{1,...,K\}$.
The local topic embedding is then: 
\begin{alignat}{2}
q_{it} &=\; \sum_{k=1}^K \alpha_{it}^k \otimes e_k 
\label{eq:local_topic_dist}
\end{alignat}

\noindent with $q_{it}\in\mathbb{R}^n$, $\alpha_{it} \in \mathbb{R}^{K}$. 
Here, $\otimes$ denotes multiplication of a vector by a scalar.

We add the local topic embedding into the GRU cell to rewrite the formulae as follows:
\begin{align}
r_t &= \sigma (W_r x_t + U_r h_{t-1} + V_r \boldsymbol{q_{t-1}}) \label{eq:GRU_r} \\
z_t &= \sigma (W_z x_t + U_z h_{t-1} + V_z \boldsymbol{q_{t-1}})\\
\hat{h}_t &=  \mbox{tanh} (W_h x_t + r_t \odot (U_h h_{t-1} + V_h \boldsymbol{q_{t-1})})  \\
h_t &= (1-z_t) \odot h_{t-1} + z_t \odot \hat{h}_t \label{eq:GRU_h} 
\end{align}
where $\sigma(\cdot)$ is the sigmoid function, all the $W$, $U$ and $V$s are weight matrices which are learned in the training process, $\odot$ denotes the element-wise product. The reset gate $r_t$ controls how much past state information is to be ignored in the current state update. The update gate $z_t$ controls how much information from the previous hidden state will be kept. The hidden state $h_t$ is computed as the interpolation between the previous state $h_{t-1}$ and the current candidate state $\hat{h}_t$.

In the above formulation, the hidden state in the current word position not only depends on the current input and the previous hidden state, but also takes into account the local topic embedding of the previous word. 
Since some of those words may be more informative than others in constituting the overall sentence meaning, we aggregate these representations with a final attention mechanism:  
\begin{alignat}{3}
v_{it}  &= \mbox{tanh}(W_v h_{it}) \label{eq:final_att1}\\
\beta_{it} &= \mbox{softmax}(v_{it}^\intercal v_{w})\\
s_{i} &= \sum_{t=1}^t \beta_{it} \otimes h_{it} 
\end{alignat}
where 
$\beta_{it}$ is the attention weight for the hidden state $h_{it}$ and $s_i \in \mathbb{R}^{n}$ is the sentence representation for the $i$th sentence.

\subsection{Sentence Encoder}
Given each sentence representation $s_i$ in document $d$ where $i=\{1,...,d_L\}$ and $d_L$ denotes the document length, we can form the document representation using the proposed topical GRU in a similar way. For each sentence $i$, its context vector is $h_{i} = [\overrightarrow{h_{i}}, \overleftarrow{h_{i}}]$, which captures the contextual information of the whole document centred at $s_i$. 

We follow an approach analogous to the topic-dependent word encoder and generate the local topic embedding for $i$th sentence:
\begin{alignat}{2}
u_{i} &= \mbox{tanh}(W_s h_{i}) &\quad W_s &\in \mathbb{R}^{n\times n} \\
\alpha_{i}^k &= \mbox{softmax}(u_{i}^\intercal e_k) &\quad k &\in \{1,...,K\}\\
q_{i} &= \sum_{k=1}^K \alpha_{i}^k \otimes e_k &\quad q_{i} &\in \mathbb{R}^n
\end{alignat}
where $q_i$ is local topic embedding for sentence $i$. We add the local topic embedding into the GRU cell as in Eq. \ref{eq:GRU_r}-\ref{eq:GRU_h}. 

Analogously to the word encoder, those sentences contribute differently to the overall document meaning; thus, we aggregate these representations with an attention mechanism similar to the final attention mechanism described in Section \ref{sec:topic_word_encoder}.

\subsection{Multi-Task Learning}
Finally, for each document $d$, we feed its representation $m_d$ into the task-specific softmax layers, each one defined as follows:
\begin{equation}
p_d = \mbox{softmax}(W_d m_d) \quad  W_d \in \mathbb{R}^{C\times n}
\end{equation}

\noindent where $C$ denotes the total number of classes. The training loss is defined as the total cross-entropy of all documents computed for each task:
\begin{equation}
L_{task} = -\sum_{d=1}^D \sum_{c=1}^C y_{d,c} \log p_{d,c}
\end{equation}
where $y_{d,c}$ is the binary indicator (0 or 1) if class label $c$ is the correct classification for document $d$.
We compute the overall loss as a weighted sum over the task-specific losses:
\begin{equation}
L_{total} = \sum_{j=1}^J \omega_{j} L(\hat{y}^{(j)}, y^{(j)})
\end{equation}
\noindent where $J$ is the number of tasks, $\omega_{j}$ is the weight for each task, $y^{(j)}$ are the ground-truth labels in task $j$ and $\hat{y}^{(j)}$ are the predicted labels in task $j$.

\subsection{Topic Extraction}  \label{ssec:topic_extraction} 

Once our model is trained, we can feed the test set and collect the local topic embedding $q_{it}$ associated to each word (Eq. \ref{eq:local_topic_dist}), collecting a set of $n$-dimensional vectors for each occurrence of words in text. This mechanism can be interpreted analogously to models generating deep contextualised word representations based on language model, where each word occurrence has a unique representation based on the context in which it appears \citep{Peters18, Devlin18}.

The local representation $q_{it}$ in our model results from the interaction with the global topic embeddings, which encode the word co-occurrence patterns characterizing the corpus. 
We posit that these vectors can give us an insight about the topic and polarity relations among words. Therefore, we first project these representations into a two-dimensional space by applying the t-SNE \citep{VanDerMaaten08}; then, the resulting word vectors are clustered by applying the \textit{K-means} algorithm.
We create a fixed number of clusters $k$, whose value is tuned by maximizing the topic coherence for $k \in [50,100,200]$. We use the distance of each word to the centroid of a topic cluster to rank words within a cluster.
Similarly, we cluster sentences based on the representation resulting from the sentence-level topical attention layer. This encoding synthesises both the main topic and polarity characterizing the sentence.

\section{Experimental Setup}

\renewcommand{\arraystretch}{0.9}
\begin{table}[!ht]
\centering
\begin{tabular}{@{}lrr@{}}
    \toprule
    \textbf{Dataset}  & \textbf{Yelp18}  & \textbf{Amazon}\\
    \midrule
    Sentiment classes      & 3 & 3 \\
    Domain categories & 5 & 5 \\
    Documents    & 75,000 &  75,000\\
    Average \#s   & 9.7 & 6.7 \\
    Average \#w   & 15.9 & 16.7 \\
    Vocabulary   & $\sim85 \times 10^3$ & $\sim100 \times 10^3$ \\
    Tokens       & $\sim11.7 \times 10^6$ & $\sim8.5 \times 10^3$ \\
    \bottomrule
\end{tabular}
\caption{Dataset statistics with \#s number of sentences per document and and \#w of words per sentence.}
\label{tb:dataset_statistics}
\end{table}

\subsection{Datasets}
\label{ss:datasets}
We gathered two balanced datasets of reviews from the publicly available Yelp Dataset Challenge dataset in 2018 and the Amazon Review Dataset\footnote{http://jmcauley.ucsd.edu/data/amazon/} \citep{McAuley15}, preserving the meta-information needed for a multi-task learning scenario. Each review is accompanied with one of the three ratings, \emph{positive}, \emph{negative} or \emph{neutral} and comes from five of the most frequent domains\footnote{For Yelp: \textit{restaurants}, \textit{shopping}, \textit{home services}, \textit{health \& medical} and \textit{automotive}. For Amazon: \textit{Pet supplies}, \textit{electronics}, \textit{health personal care}, \textit{clothes shoes} and \textit{home and kitchen}.}. 
Those ratings are the human labeled review scores regarded as gold standard sentiment labels during the experimentation. For each pair of domain and rating, we randomly sample 3,000 reviews, collecting a total of 75,000 reviews. 
To make it possible for others to replicate our results, we make both the dataset and our source code publicly available\footnote{\url{https://github.com/gabrer/topic_dependent_attention_model}}. 
Table \ref{tb:dataset_statistics} summarizes the statistics of the datasets.

\subsection{Baselines}
\label{ss:baselines}
We train our proposed TDAM with multi-task learning to perform sentiment and domain classification simultaneously. 
We compare the performance of TDAM with the following baselines on both sentiment classification and topic extraction:

\begin{itemize}
\item \textbf{BiLSTM} \citep{Hochreiter97} or \textbf{BiGRU} \citep{Cho14}: Both models consider a whole document as a single text sequence. The average of the hidden states is used as features for classification. 

\item \textbf{Hierarchical Attention Network (HAN)} \citep{Yang16}: The hierarchical structure of this attention model learns word and sentence representations through two additive attention levels. 

\item \textbf{Supervised-LDA (S-LDA)} \citep{Blei08}:  It builds on top of the latent Dirichlet allocation (LDA) \citep{Blei03} adding a response variable associated with each document (e.g. review's rating or category).

\item \textbf{\textsc{Scholar}} \citep{Card18}: A neural framework for topic models with metadata incorporation without the need of deriving model-specific inference. When metadata are labels, the model infers topics that are relevant to those labels.

\end{itemize}

The baselines, such as BiLSTM, BiGRU and HAN, are additionally trained with multi-task learning, similar to the setup of our model.

\subsection{Parameter Settings}
For our experiments, we split the dataset into training, development and test set in the proportion of 80/10/10 and average all the results over 5-fold cross-validation.
We perform tokenization and sentence splitting with SpaCy\footnote{https://spacy.io/}.
We do not filter any words from the dataset during the training phase; although we use the default preprocessing for models like S-LDA and \textsc{Scholar}.
Word embeddings are initialized with 200-dimensional GloVe vectors \citep{Pennington14}. 
We tune the models' hyperparameters on the development set via a grid search over combinations of learning rate $\lambda \in [0.01, 0.1]$, dropout $\delta \in [0,0.6]$ and topic vector's size $\gamma_t \in [50,200]$. Matrices are randomly initialized to be semi-orthogonal matrix \citep{SaxeMG13}; all the remaining parameters are randomly sampled from a uniform distribution in $[-0.1,0.1]$. We adopt Adam optimizer \citep{Kingma2014} and use batch size of 64, sorting documents by length (i.e. number of sentences) to accelerate training convergence; we also apply batch normalization as additional regulariser \citep{Cooijmans17}.

Once the model is trained, we extract the local topic embedding for each word occurrence in text as its contextualized word representation. These vectors are then projected to a lower-dimensional space by means of a multi-core implementation of a Tree-Based algorithm for accelerating t-SNE\footnote{https://github.com/DmitryUlyanov/Multicore-TSNE} \citep{VanDerMaaten14}. Then, we cluster these words with K-means\footnote{http://scikit-learn.org/stable/modules/clustering.html}.

\section{Evaluation and results}
\label{s:acc_results}
We report and discuss the experimental results obtained on three evaluation tasks, sentiment classification topic extraction and sentence cluster extraction. 

\renewcommand{\arraystretch}{0.9}
\begin{table}[!ht]
\centering
\begin{tabular}{@{}lcc@{}}
    \toprule
    \textbf{Methods}  & \textbf{Yelp 18} & \textbf{Amazon}  \\
    \midrule
    BiLSTM     	                &  $74.5\pm0.2$ &  $72.1\pm0.2$ \\
    BiLSTM - Mtl $\quad\quad$   &  $74.2\pm0.2$ &  $71.8\pm0.1$ \\
    BiGRU 	                    &  $75.5\pm0.1$ &  $72.5\pm0.3$ \\
    BiGRU - Mtl	    	        &  $75.4\pm0.2$ &  $72.1\pm0.3$ \\
    HAN	    	                &  $83.7\pm0.2$ &  $78.4\pm0.2$ \\
    HAN - Mtl		    	    &  $83.6\pm0.3$ &  $78.2\pm0.3$ \\
    \midrule
    S-LDA     		            &  $70.8\pm0.2$ &  $64.6\pm0.1$ \\
    \textsc{Scholar}            &  $77.3\pm0.2$ &  $71.4\pm0.2$  \\
    \midrule   
    TDAM 		                & $84.2\pm0.2$  &  $78.9\pm0.2$  \\
    TDAM - Mtl		            & $\mathbf{84.5}\pm0.3$ &  $\mathbf{79.1}\pm0.2$ \\
    \bottomrule
  \end{tabular}
\caption{Sentiment classification accuracy and standard deviation over the 5-fold cross validation.}
\label{tb:accuracy}
\end{table}

\subsection{Sentiment Classification}

We train the models under two different settings: a single and a multi-task learning scenario, where we optimize over the only review polarity or over the combination of polarity and domain, respectively.
For the latter, we denote the results with `-Mtl' in Table \ref{tb:accuracy}. 

It can be observed from the table that BiLSTM and BiGRU perform similarly. 
With hierarchical attention mechanism at both the word level and the sentence level, HAN boosts the performance by nearly 10\% on Yelp and 6\% on Amazon compared to BiLSTM and BiGRU.  
For the neural topic modeling approaches, \textsc{Scholar} outperforms traditional S-LDA by a large margin. However, \textsc{Scholar} is still inferior to HAN. With our proposed topical attentions incorporated into the hierarchical network structure, TDAM further improves on HAN. Multi-task learning does not seem to bring any benefit to sentiment classification for baseline models, though it further improves the performance of TDAM slightly.

\subsection{Effectiveness of Topical Attention}
\label{ss:ablation_study}

If we remove the topical attention and substitute our modified GRU with standard GRU, then the resulting architecture is similar to HAN \citep{Yang16} for a multi-task learning setting. In this section, we visualize the attention weights learned by HAN and TDAM to compare their results. Examples are shown in Figure \ref{fig:att_weights}.
In TDAM, topical words such as \textit{dentist} or the dentist's name, \textit{Rebecca}, are regarded as relevant by the model. Along with them, it focuses on words bearing a strong sentiment, such as \textit{nicest} or \textit{happy}. These weights are compared with the attention weights learned by the HAN, showing that it primarily focuses sentiment words and overlooks other topical words, such as \textit{dentist}.

\renewcommand{\arraystretch}{0.9}
\begin{table}[!t]
\small
\centering

\begin{tabular}{@{}rrrrrcrrrrcrrrrrrr@{}}
    \toprule
& \multicolumn{3}{c}{\textbf{Yelp18}}  & \multicolumn{3}{c}{\textbf{Amazon}} \\
$Topics=$ & $50$ & $100$ & $200$ & $50$ & $100$ & $200$ \\ 
    \midrule
       HAN   		        &  -7.22        &  -7.05 	&  -7.08        &  -13.21        &  -13.15 	&  -13.14   \\
       HAN - Mtl            &  -7.04        &  -6.94 	&  -6.93        &  -12.72        &  -12.20 	&  -12.29   \\
       
      S-LDA 		            &  -6.26        & -6.13 	& -6.15          & -9.57        & -9.41 &    -9.28 \\
    \textsc{Scholar} 	    &  -6.24        &  -6.08	&  -6.11                     &  -9.52        &  -9.46 &  -9.48      \\
   
    \textsc{Scholar-R}    &  \textbf{-6.19}    &  -6.11	&  -6.08    &  -9.34        &  \textbf{-9.09} 	&  -9.17 \\

    TDAM	            &  -6.41      &  -6.12      &  -6.09        &  -9.62        &  -9.50 	&  -9.46\\
    TDAM - Mtl          &  -6.22      &  \textbf{-6.05} &  \textbf{-5.93} & \textbf{-9.23}   &  -9.12 	&  \textbf{-9.01} \\

    \bottomrule
  \end{tabular}
\caption{Topic coherence for different number of topics. The higher the better.}
\label{tb:topic_coherence}
\end{table}

\subsection{Topic Coherence Evaluation}
\label{ss:topic_evaluation}

Among the baselines, S-LDA and \textsc{Scholar} are topic modeling methods and therefore the can directly output topics from text. In addition, we can follow the topic extraction procedure described in Section \ref{ssec:topic_extraction} to extract topics from HAN to gain an insight into the learned representations. We thus compare the topic extraction results of TDAM with these three models. Also, as previously shown in \citep{Card18}, higher regularisation on \textsc{Scholar} produced better topics. Therefore, we also report the results using \textsc{Scholar} with higher regularization, named as \textsc{Scholar-R}.

To evaluate the quality of topics, we use the topic coherence measure\footnote{https://github.com/dice-group/Palmetto} proposed in \citep{Roder15} 
which has been shown outperforming all the other existing topic coherence measures. We can observe from Table \ref{tb:topic_coherence} that HAN gives the worse topic coherence results, showing that simply extracting topics using the attention weights is not feasible. With the incorporation of domain category information through multi-task learning, HAN-Mtl gives slightly better coherence results. Among topic modeling approaches, \textsc{Scholar-R} with higher regularization generates more coherence topics compared to \textsc{Scholar}, which outperforms S-LDA. TDAM gives similar topic coherence results as \textsc{Scholar-R} on some topic numbers. TDAM-Mtl improves over TDAM and generates the best coherence results on 2 out of 3 topic settings for both Yelp18 and Amazon, showing higher coherence scores overall.

\subsection{Aspect-Polarity Coherence Evaluation}
\label{ss:phrase_clustering}
To assess the effectiveness of our proposed TDAM in extracting polarity-bearing topics, we use the annotated dataset provided in the SemEval 2016 Task 5 for aspect-based sentiment analysis\footnote{http://alt.qcri.org/semeval2016/task5/}; this provides sentence-level annotations about different aspects (e.g. {\small \texttt{FOOD\#QUALITY}}) and polarities ({\small \texttt{pos}, \texttt{neut}, \texttt{neg}}) in restaurant and laptop reviews. 

We join the training set of restaurant and laptop reviews with the Yelp18 and Amazon dataset, respectively. With the same approach adopted for topic extraction, we use the test sets to generate sentence clusters and evaluate their \textit{aspect-polarity coherence}, defined as the ratio of sentences sharing a common aspect and sentiment in a cluster. For the two topic modeling approaches, S-LDA and \textsc{Scholar}, we generate sentence clusters based on the generative probabilities of sentences conditional on topics.
Note that although the SemEval dataset provides the sentence-level annotations of aspects and polarities, these were NOT used for the training of the models here. We only use the gold standard annotations of aspects and polarities in the test set to evaluate the quality of the extracted polarity-bearing topics. 

We generate multiple clusters, i.e. (50,100,150), representing polarity-bearing aspects and report the results in Table \ref{tb:aspect_coherence}, which shows the ratio of sentence clusters with more than \textit{threshold} sentences sharing a common aspect (values in brackets) or a common aspect-polarity. 
We can observe that the topic modeling approaches struggle in generating coherent aspect-polarity clusters with at least $50\%$ of common aspect-polarities. The two hierarchical models, HAN and TDAM, have significantly more coherent aspect-polarity clusters compared to S-LDA and \textsc{Scholar}, and both benefit from multi-task learning. For all the models, results on SemEval-Restaurant are better than those obtained on SemEval-Laptop. This might be partly attributed to the abundant restaurant reviews on Yelp18 compared to the laptop-related reviews on Amazon. Overall, TDAM-Mtl gives the best results. 

We also show some example sentence clusters produced by HAN and TDAM under multi-task learning in Table \ref{tb:aspect_examples}. HAN discriminates rather effectively positive sentences (the majority in the cluster) from negative and neutral ones. However, despite several sentences sharing the same polarity, their topics/aspects are quite heterogeneous. TADM phrases are rather coherent overall, both in terms of topics and expressed sentiment.

These results are encouraging. Our TDAM is able to detect coherent aspects and also polarity-bearing aspects despite using no aspect-level annotations at all. Considering it is very time consuming to provide aspect-level annotations, TDAM could be used to bootstrap the training of aspect-based sentiment detectors. 

\section{Conclusion}
\noindent We have presented a new topic-dependent attention model for sentiment classification and topic extraction. The conjunction of topical recurrent unit and multi-task learning framework has been shown to be an effective combination to generate representations for more accurate sentiment classification, meaningful topics and for side task of polarity-bearing aspects detection. In future, we will extend the model to deal with discourse-level sentiments \citep{Feng12}.

\begin{sidewaystable} 
\centering
\resizebox{1\columnwidth}{!}{%
\begin{tabular}{@{}rcccccc|ccccc@{}}
\toprule
\textbf{Methods} &  \textbf{Topics} & \multicolumn{3}{r}{\textbf{SemEval-Restaurant}} & \multicolumn{5}{r}{\textbf{SemEval-Laptop}} \\ 
    \midrule
     & $threshold$  & $\ge 50\%$ & $\ge 60\%$ & $\ge 70\%$ & $\ge 80\%$ & $\ge 90\%$ & $\ge 50\%$ & $\ge 60\%$ & $\ge 70\%$ & $\ge 80\%$ & $\ge 90\%$ \\ 
     \cmidrule{2-12}
                        &    50       & (0.52) 0.40    & (0.28) 0.24   &   (0.14) 0.10   & (0.04)  0.02   &  (0.00) 0.00   & (0.18) 0.15    & (0.15) 0.12   &   (0.08) 0.03   &  (0.03) 0.01   & (0.00)  0.00  \\
       HAN   	        &	 100      & (0.64) 0.47    & (0.36) 0.27   &   (0.14) 0.11   & (0.09) 0.07   &  (0.03) 0.03    & (0.28) 0.26    & (0.21) 0.20   &   (0.12) 0.08   &  (0.05) 0.04   &  (0.01) 0.01 \\
                        &    150      & (0.70) 0.59    & (0.39) 0.32   &   (0.21) 0.16   &  (0.14) 0.12   &  (0.12) 0.08   & (0.37) 0.34    & (0.28) 0.23   &   (0.14) 0.10   &  (0.8) 0.07   &  (0.04) 0.03 \\
    \cmidrule{2-12}
                        &    50       & (0.56) 0.40    & (0.36) 0.28   &   (0.26) 0.18   &  (0.12) 0.10   &  (0.06) 0.02    & (0.24) 0.19    & (0.18) 0.15   &  (0.12) 0.04   &  (0.04) 0.02   &  (0.03) 0.01 \\
       HAN-Mtl      	&	 100      & (0.64) 0.52    & (0.51) 0.40   &   (0.26) 0.22   &  (0.17) 0.13   &  (0.10) 0.06    & (0.31) 0.27    & (0.26) 0.19   &   (0.16) 0.09   &  (0.08) 0.04   &  (0.04) 0.03  \\
                        &    150      & (0.72) 0.63    & (0.51) 0.43   &   (0.30) 0.22   &  (\textbf{0.21}) 0.12   &  (0.14) 0.08    & (0.41) 0.38    & (0.35) 0.24   &   (0.17) 0.12   &  (0.12) 0.08   &  (0.05) 0.03 \\
    \cmidrule{2-12}
     		              &    50       & (0.18) 0.12    & (0.08) 0.03   &   (0.02) 0.01   &  (0.00) 0.00   &  (0.00) 0.00    & (0.09) 0.07    & (0.08) 0.05   &   (0.03) 0.01   & (0.02)  0.00   & (0.00)  0.00  \\
    \textsc{S-LDA} 	      &    100      & (0.24) 0.21    & (0.11) 0.10   &   (0.03) 0.02   &  (0.00) 0.00   &  (0.00) 0.00    & (0.15) 0.14    & (0.10) 0.06   &   (0.04) 0.01   & (0.01)  0.00   & (0.00)  0.00   \\
    		              &    150      & (0.39) 0.35    & (0.19) 0.16   &   (0.05) 0.04   &  (0.01) 0.01   &  (0.01) 0.01    & (0.27) 0.24    & (0.14) 0.11   &   (0.08) 0.04   & (0.2)  0.02   &  (0.00) 0.00  \\
    \cmidrule{2-12}
     		              &    50       & (0.31) 0.18    & (0.22) 0.10   &  (0.04) 0.03   &  (0.01) 0.01   &  (0.00) 0.00     & (0.14) 0.10    & (0.08) 0.04   &  (0.06) 0.02   & (0.03) 0.01   &  (0.00) 0.00  \\
    \textsc{Scholar-R} 	  &    100      & (0.39) 0.24    & (0.24) 0.13   &  (0.08) 0.04   &  (0.03) 0.01   &  (0.01) 0.01     & (0.21) 0.15    & (0.16) 0.08   &  (0.09) 0.05   &  (0.03) 0.01   &  (0.01) 0.01  \\
    		              &    150      & (0.43) 0.36    & (0.28) 0.19   &  (0.11) 0.10   &  (0.04) 0.03   &  (0.01) 0.01     & (0.34) 0.26    & (0.21) 0.13   &  (0.12) 0.06   &  (0.05) 0.02   &  (0.02) 0.01  \\
    \cmidrule{2-12}
     		              &    50       & (0.54) 0.42    & (0.30) 0.24   &  (0.12) 0.08   &  (0.06) 0.04   & (0.02) 0.00                & (0.19) 0.17    & (0.15) 0.14  & (0.09)  0.06   & (0.03)  0.02   &  (0.00) 0.00 \\
    \textsc{TDAM} 		  &    100      & (0.63) 0.55   & (0.40) 0.31   &   (0.21) 0.16   &  (0.14) 0.10   &  (0.10) 0.06               & (0.38) 0.29    & (0.24)  0.18  & (0.12)   0.10  & (0.05)  0.04   &  (0.02) 0.02\\
    		              &    150      & (0.73) 0.65    & (0.43) 0.36   &   (0.28) \textbf{0.26}   & (0.19) 0.15   &  (\textbf{0.16}) 0.13      & (0.39) 0.37    &  (0.28) 0.25  &  (0.16) 0.13   &  (0.08) 0.08   &  (0.05) 0.03\\
    \cmidrule{2-12}
     		              &    50      & (0.68) 0.51    & (\textbf{0.52}) 0.38   &  (0.24) 0.20   & (0.14) 0.12   &  (0.06) 0.04   & (0.31) 0.25         & (0.26) 0.17   & (0.22)  0.13   & (0.13) 0.07   & (0.04) 0.02 \\
    \textsc{TDAM}-Mtl    &    100      & (0.72) 0.58    & (0.47) 0.39   &  (0.31) 0.24   &  (0.19)0.16   &  (0.13) 0.12  & (0.39) 0.32          & (0.38) 0.24   &  (0.26) 0.15   & (0.12) 0.09   & (0.02) 0.0\\
    		             &    150      & (\textbf{0.80}) \textbf{0.68}    & (0.50) \textbf{0.40}   &  (\textbf{0.32}) 0.25   &  (0.20) \textbf{0.16}    &  (\textbf{0.16}) \textbf{0.14}   & (\textbf{0.48}) \textbf{0.43}    & (\textbf{0.42}) \textbf{0.31}   &  (\textbf{0.26}) \textbf{0.18}   &  (\textbf{0.17}) \textbf{0.11}   &  (\textbf{0.09}) \textbf{0.05}   \\
     	   
    \bottomrule
  \end{tabular}
  }
\caption{Ratio of clusters where at least $x\%$ sentences sharing the same aspect (values in brackets) and sharing the same aspect-polarity (i.e. both aspect and polarity are correct).}
\label{tb:aspect_coherence}
\end{sidewaystable}

\begin{sidewaystable} 
\centering
\resizebox{1\textwidth}{!}{ 
\scriptsize
\centering
\begin{tabular}{@{}ll@{}}
    \textbf{HAN} & \textbf{TDAM} \\
    \midrule
    \multicolumn{2}{c}{\textbf{Positive polarity - Food\#Quality}}\\
    \midrule
    
    \begin{tabular}[x]{@{}l@{}}1) wait the half hour with a cup of joe ,\\ and enjoy more than your average breakfast .\end{tabular} \hfill	\texttt{FOOD\#QUALITY	pos}  & 1) the food was all good but it was way too	\hfill \texttt{FOOD\#QUALITY	neg} \\
    
     2) space was limited , but the food made up for it .	\hfill \texttt{RESTAURANT\#MISCELLANEOUS	neg} & 2) the pizza 's are light and scrumptious .	\hfill \texttt{FOOD\text{\#}STYLE\_OPTIONS	pos} \\
    
     3)the prices should have been lower .\hfill \texttt{FOOD\#STYLE\_OPTIONS	neg}  & 3) the food is great and they make a mean bloody mary .	\hfill \texttt{FOOD\#QUALITY	pos} \\
    
    4) the crowd is mixed yuppies , young and old .	\hfill \texttt{RESTAURANT\#MISCELLANEOUS	neut} & 4) great draft and bottle selection and the pizza rocks .	\hfill \texttt{FOOD\#QUALITY	pos} \\
    
     \begin{tabular}[x]{@{}l@{}}  5)making the cakes myself since i was about seven  - but \\something about these little devils gets better every time .\end{tabular}	\hfill \texttt{FOOD\#QUALITY	pos} & 5) the food is simply unforgettable !	\hfill \texttt{FOOD\#QUALITY	pos}\\
    
    6) mmm ... good !	\hfill \texttt{RESTAURANT\#GENERAL	pos}  & 6)the food is great, the bartenders go that extra mile.	\hfill \texttt{FOOD\#QUALITY	pos} \\
    
    7) the service is so efficient you can be in and out of there quickly .	\hfill \texttt{SERVICE\#GENERAL	pos} &  7)the food is sinful.	\hfill \texttt{FOOD\#QUALITY	pos}\\
    
    8) service was decent .	\hfill \texttt{SERVICE\#GENERAL	neut}  & 8)the sushi here is delicious !	\hfill \texttt{FOOD\#QUALITY	pos} \\
    
    9) their specialty rolls are impressive \hfill \texttt{FOOD\#QUALITY	pos} & 9) the food was great !	\hfill \texttt{FOOD\#QUALITY	pos} \\
    
    \begin{tabular}[x]{@{}l@{}} 10) it was n\'t the freshest seafood ever , but the taste \\ and presentation was ok .\end{tabular} 	\hfill \texttt{FOOD\#STYLE\_OPTIONS	neut}  & 10) good eats .	\hfill \texttt{FOOD\#QUALITY	pos} \\
    
    \midrule
    \multicolumn{2}{c}{\textbf{Negative polarity - Food\#Quality}}\\
    \midrule

    \begin{tabular}[x]{@{}l@{}}1) the pancakes were certainly inventive but \$ 8.50 \\ for 3 - 6 " pancakes ( one of them was more like 5 " )	\end{tabular} \hfill \texttt{FOOD\#STYLE\_OPTIONS	neg}  & 1) i may not be a sushi guru	\hfill \texttt{FOOD\#QUALITY	neg} \\
    
    \begin{tabular}[x]{@{}l@{}}2)a beautiful assortment of enormous white gulf prawns , \\smoked albacore tuna, [..] and a tiny pile of dungeness \end{tabular} 	\hfill \texttt{FOOD\#STYLE\_OPTIONS	pos} & 2) rice is too dry , tuna was n't so fresh either .	\hfill \texttt{FOOD\#QUALITY	neg} \\
    
    3) space was limited , but the food made up for it .	\hfill \texttt{RESTAURANT\#MISCELLANEOUS	neg} & 3)\begin{tabular}[x]{@{}l@{}}the only way this place survives with such average food is \\ because most customers are one-time customer tourists\end{tabular}\hfill \texttt{FOOD\#QUALITY	neg}\\
    
     4) the portions are big though , so do not order too much .  \hfill \texttt{FOOD\#STYLE\_OPTIONS	neut} & 4) the portions are big though , so do not order too much .	\hfill \texttt{FOOD\#STYLE\_OPTIONS	neut}\\
    
    5) not the biggest portions but adequate .	\hfill \texttt{FOOD\#QUALITY	pos} & \begin{tabular}[x]{@{}l@{}} 5) the only drawback is that this place is really expensive\\ and the portions are on the small side .\end{tabular}	\hfill \texttt{RESTAURANT\#PRICES	neg} \\
   
    \begin{tabular}[x]{@{}l@{}} 6)the waiter was a bit unfriendly and the feel of the \\ restaurant was crowded .\end{tabular} 	\hfill \texttt{SERVICE\#GENERAL	neg} &  6)\begin{tabular}[x]{@{}l@{}}  but i can tell you that the food here is just okay \\and that there is not much else to it .\end{tabular} 	\hfill \texttt{FOOD\#QUALITY	neg} \\
   
   7) food was fine , with a some little - tastier - than - normal salsa .	\hfill \texttt{FOOD\#QUALITY	pos} & 7) and they give good quantity for the price .	\hfill \texttt{FOOD\#STYLE\_OPTIONS	pos} \\
   
   8) i got the shellfish and shrimp appetizer and it was alright .	\hfill \texttt{FOOD\#QUALITY	neut} & 8) food was fine , with a some little - tastier - than - normal salsa .	\hfill \texttt{FOOD\#QUALITY	pos} \\
   
     9) once seated it took about 30 minutes to finally get the meal .	\hfill \texttt{FOOD\#GENERAL	neg} & \begin{tabular}[x]{@{}l@{}} 9)our drinks kept coming but our server came by \\ a couple times .  \end{tabular} 	\hfill \texttt{SERVICE\#GENERAL	pos} \\
   
     \begin{tabular}[x]{@{}l@{}} 10) the food is here is incredible , though the quality is inconsistent\\ during lunch .	 \end{tabular}  \hfill \texttt{FOOD\#QUALITY	pos} & 10) nice food but no spice !	\hfill \texttt{FOOD\#QUALITY	neg} \\
     
\bottomrule
\end{tabular}
}
\caption{Clusters of positive and negative aspects about {\small\texttt{FOOD\#QUALITY}} experiences from SemEval16. On the left sentences clustered with HAN, on the right the ones clustered with TDAM. The aspect and polarity label for each sentence are the gold standard annotations.}
\label{tb:aspect_examples}

\end{sidewaystable}

\bibliography{tdam}

\begin{thebibliography}{41}
\expandafter\ifx\csname natexlab\endcsname\relax\def\natexlab#1{#1}\fi
\providecommand{\url}[1]{\texttt{#1}}
\providecommand{\href}[2]{#2}
\providecommand{\path}[1]{#1}
\providecommand{\DOIprefix}{doi:}
\providecommand{\ArXivprefix}{arXiv:}
\providecommand{\URLprefix}{URL: }
\providecommand{\Pubmedprefix}{pmid:}
\providecommand{\doi}[1]{\href{http://dx.doi.org/#1}{\path{#1}}}
\providecommand{\Pubmed}[1]{\href{pmid:#1}{\path{#1}}}
\providecommand{\bibinfo}[2]{#2}
\ifx\xfnm\relax \def\xfnm[#1]{\unskip,\space#1}\fi
\bibitem[{Abdi et~al.(2019)Abdi, Shamsuddin, Hasan \& Piran}]{Abdi19}
\bibinfo{author}{Abdi, A.}, \bibinfo{author}{Shamsuddin, S.~M.},
  \bibinfo{author}{Hasan, S.}, \& \bibinfo{author}{Piran, J.}
  (\bibinfo{year}{2019}).
\newblock \bibinfo{title}{Deep learning-based sentiment classification of
  evaluative text based on multi-feature fusion}.
\newblock {\it \bibinfo{journal}{Information Processing \& Management}\/},
  {\it \bibinfo{volume}{56}\/}, \bibinfo{pages}{1245 -- 1259}.
\bibitem[{Bahdanau et~al.(2015)Bahdanau, Cho \& Bengio}]{Bahdanau2014}
\bibinfo{author}{Bahdanau, D.}, \bibinfo{author}{Cho, K.}, \&
  \bibinfo{author}{Bengio, Y.} (\bibinfo{year}{2015}).
\newblock \bibinfo{title}{Neural machine translation by jointly learning to
  align and translate}.
\newblock In {\it \bibinfo{booktitle}{3rd International Conference on Learning
  Representations, {ICLR} 2015, San Diego, CA, USA}\/}.
\bibitem[{Bengio(2017)}]{Bengio17-CPrior}
\bibinfo{author}{Bengio, Y.} (\bibinfo{year}{2017}).
\newblock \bibinfo{title}{The consciousness prior}.
\newblock {\it \bibinfo{journal}{CoRR}\/},  {\it
  \bibinfo{volume}{abs/1709.08568}\/}.
\bibitem[{Blei et~al.(2003)Blei, Ng \& Jordan}]{Blei03}
\bibinfo{author}{Blei, D.~M.}, \bibinfo{author}{Ng, A.~Y.}, \&
  \bibinfo{author}{Jordan, M.~I.} (\bibinfo{year}{2003}).
\newblock \bibinfo{title}{Latent {D}irichlet {A}llocation}.
\newblock {\it \bibinfo{journal}{Journal of Machine Learning Research}\/},
  {\it \bibinfo{volume}{3}\/}, \bibinfo{pages}{993--1022}.
\bibitem[{Card et~al.(2018)Card, Tan \& Smith}]{Card18}
\bibinfo{author}{Card, D.}, \bibinfo{author}{Tan, C.}, \&
  \bibinfo{author}{Smith, N.~A.} (\bibinfo{year}{2018}).
\newblock \bibinfo{title}{Neural {M}odels for {D}ocuments with {M}etadata}.
\newblock In {\it \bibinfo{booktitle}{Proceedings of the 56th Annual Meeting of
  the Association for Computational Linguistics, ACL 2018}\/} (pp.
  \bibinfo{pages}{2031--2040}).
\newblock \bibinfo{address}{Melbourne, Australia}.
\bibitem[{Chen et~al.(2016)Chen, Sun, Tu, Lin \& Liu}]{Chen16}
\bibinfo{author}{Chen, H.}, \bibinfo{author}{Sun, M.}, \bibinfo{author}{Tu,
  C.}, \bibinfo{author}{Lin, Y.}, \& \bibinfo{author}{Liu, Z.}
  (\bibinfo{year}{2016}).
\newblock \bibinfo{title}{Neural sentiment classification with user and product
  attention}.
\newblock In {\it \bibinfo{booktitle}{Proceedings of the 2016 Conference on
  Empirical Methods in Natural Language Processing, EMNLP 2016}\/} (pp.
  \bibinfo{pages}{1650--1659}).
\newblock \bibinfo{address}{Austin, Texas, USA.}
\bibitem[{Chen \& Cardie(2018)}]{Chen18}
\bibinfo{author}{Chen, X.}, \& \bibinfo{author}{Cardie, C.}
  (\bibinfo{year}{2018}).
\newblock \bibinfo{title}{Multinomial adversarial networks for multi-domain
  text classification}.
\newblock In {\it \bibinfo{booktitle}{Proceedings of the 2018 Conference of the
  North {A}merican Chapter of the Association for Computational Linguistics:
  Human Language Technologies, NAACL 2018}\/} (pp.
  \bibinfo{pages}{1226--1240}).
\newblock \bibinfo{address}{New Orleans, Louisiana}.
\bibitem[{Cho et~al.(2014)Cho, van Merri{\"{e}}nboer, G{\"{u}}l{\c c}ehre,
  Bahdanau, Bougares, Schwenk \& Bengio}]{Cho14}
\bibinfo{author}{Cho, K.}, \bibinfo{author}{van Merri{\"{e}}nboer, B.},
  \bibinfo{author}{G{\"{u}}l{\c c}ehre, {\c C}.}, \bibinfo{author}{Bahdanau,
  D.}, \bibinfo{author}{Bougares, F.}, \bibinfo{author}{Schwenk, H.}, \&
  \bibinfo{author}{Bengio, Y.} (\bibinfo{year}{2014}).
\newblock \bibinfo{title}{Learning phrase representations using rnn
  encoder--decoder for statistical machine translation}.
\newblock In {\it \bibinfo{booktitle}{Proceedings of the 2014 Conference on
  Empirical Methods in Natural Language Processing, EMNLP 2014}\/} (pp.
  \bibinfo{pages}{1724--1734}).
\newblock \bibinfo{address}{Doha, Qatar}.
\bibitem[{Cooijmans et~al.(2017)Cooijmans, Ballas, Laurent, G{\"{u}}l{\c c}ehre
  \& Courville}]{Cooijmans17}
\bibinfo{author}{Cooijmans, T.}, \bibinfo{author}{Ballas, N.},
  \bibinfo{author}{Laurent, C.}, \bibinfo{author}{G{\"{u}}l{\c c}ehre, {\c
  C}.}, \& \bibinfo{author}{Courville, A.} (\bibinfo{year}{2017}).
\newblock \bibinfo{title}{Recurrent batch normalization}.
\newblock In {\it \bibinfo{booktitle}{Proceedings of the 2017 International
  Conference for Learning Representations, ICLR 2017}\/}.
\newblock \bibinfo{address}{Touloun, France}.
\bibitem[{Devlin et~al.(2019)Devlin, Chang, Lee \& Toutanova}]{Devlin18}
\bibinfo{author}{Devlin, J.}, \bibinfo{author}{Chang, M.},
  \bibinfo{author}{Lee, K.}, \& \bibinfo{author}{Toutanova, K.}
  (\bibinfo{year}{2019}).
\newblock \bibinfo{title}{{BERT:} pre-training of deep bidirectional
  transformers for language understanding}.
\newblock In {\it \bibinfo{booktitle}{Proceedings of the 2019 Conference of the
  North American Chapter of the Association for Computational Linguistics:
  Human Language Technologies, {NAACL-HLT} 2019}\/} (pp.
  \bibinfo{pages}{4171--4186}).
\newblock \bibinfo{address}{Minneapolis, USA.}
\bibitem[{Dieng et~al.(2017)Dieng, Wang, Gao \& Paisley}]{Dieng16}
\bibinfo{author}{Dieng, A.~B.}, \bibinfo{author}{Wang, C.},
  \bibinfo{author}{Gao, J.}, \& \bibinfo{author}{Paisley, J.~W.}
  (\bibinfo{year}{2017}).
\newblock \bibinfo{title}{Topic{RNN}: {A} recurrent neural network with
  long-range semantic dependency}.
\newblock In {\it \bibinfo{booktitle}{Proceedings of the 2017 International
  Conference for Learning Representations, ICLR 2017}\/}.
\newblock \bibinfo{address}{Touloun, France}.
\bibitem[{Feng \& Hirst(2012)}]{Feng12}
\bibinfo{author}{Feng, V.~W.}, \& \bibinfo{author}{Hirst, G.}
  (\bibinfo{year}{2012}).
\newblock \bibinfo{title}{Text-level discourse parsing with rich linguistic
  features}.
\newblock In {\it \bibinfo{booktitle}{Proceedings of the 50th Annual Meeting of
  the Association for Computational Linguistics, ACL 2012}\/} (pp.
  \bibinfo{pages}{60--68}).
\newblock \bibinfo{address}{Jeju Island, Korea}.
\bibitem[{Hermann et~al.(2015)Hermann, Kocisky, Grefenstette, Espeholt, Kay,
  Suleyman \& Blunsom}]{hermann2015teaching}
\bibinfo{author}{Hermann, K.~M.}, \bibinfo{author}{Kocisky, T.},
  \bibinfo{author}{Grefenstette, E.}, \bibinfo{author}{Espeholt, L.},
  \bibinfo{author}{Kay, W.}, \bibinfo{author}{Suleyman, M.}, \&
  \bibinfo{author}{Blunsom, P.} (\bibinfo{year}{2015}).
\newblock \bibinfo{title}{Teaching machines to read and comprehend}.
\newblock In {\it \bibinfo{booktitle}{Advances in Neural Information Processing
  Systems 28, NIPS 2015}\/} (pp. \bibinfo{pages}{1693--1701}).
\newblock \bibinfo{address}{Montreal, Canada}.
\bibitem[{Hochreiter \& Schmidhuber(1997)}]{Hochreiter97}
\bibinfo{author}{Hochreiter, S.}, \& \bibinfo{author}{Schmidhuber, J.}
  (\bibinfo{year}{1997}).
\newblock \bibinfo{title}{Long {S}hort-{T}erm {M}emory}.
\newblock {\it \bibinfo{journal}{Neural {C}omputation}\/},  {\it
  \bibinfo{volume}{9}\/}, \bibinfo{pages}{1735--1780}.
\bibitem[{Jin et~al.(2018)Jin, Luo, Zhu \& Zhuo}]{Jin18}
\bibinfo{author}{Jin, M.}, \bibinfo{author}{Luo, X.}, \bibinfo{author}{Zhu,
  H.}, \& \bibinfo{author}{Zhuo, H.~H.} (\bibinfo{year}{2018}).
\newblock \bibinfo{title}{Combining deep learning and topic modeling for review
  understanding in context-aware recommendation}.
\newblock In {\it \bibinfo{booktitle}{Proceedings of the 2018 Conference of the
  North American Chapter of the Association for Computational Linguistics:
  Human Language Technologies, NAACL 2018}\/} (pp.
  \bibinfo{pages}{1605--1614}).
\newblock \bibinfo{address}{New Orleans, Louisiana}.
\bibitem[{Kastrati et~al.(2019)Kastrati, Imran \& Yayilgan}]{Kastrati19}
\bibinfo{author}{Kastrati, Z.}, \bibinfo{author}{Imran, A.~S.}, \&
  \bibinfo{author}{Yayilgan, S.~Y.} (\bibinfo{year}{2019}).
\newblock \bibinfo{title}{The impact of deep learning on document
  classification using semantically rich representations}.
\newblock {\it \bibinfo{journal}{Information Processing \& Management}\/},
  {\it \bibinfo{volume}{56}\/}, \bibinfo{pages}{1618 -- 1632}.
\bibitem[{Kingma \& Ba(2015)}]{Kingma2014}
\bibinfo{author}{Kingma, D.~P.}, \& \bibinfo{author}{Ba, J.}
  (\bibinfo{year}{2015}).
\newblock \bibinfo{title}{Adam: A method for stochastic optimization}.
\newblock In {\it \bibinfo{booktitle}{Proceedings of the 2015 International
  Conference for Learning Representations, ICLR 2015}\/}.
\newblock \bibinfo{address}{San Diego, USA}.
\bibitem[{Liu et~al.(2016)Liu, Qiu \& Huang}]{Liu16}
\bibinfo{author}{Liu, P.}, \bibinfo{author}{Qiu, X.}, \&
  \bibinfo{author}{Huang, X.} (\bibinfo{year}{2016}).
\newblock \bibinfo{title}{Deep multi-task learning with shared memory for text
  classification}.
\newblock In {\it \bibinfo{booktitle}{Proceedings of the 2016 Conference on
  Empirical Methods in Natural Language Processing, {EMNLP} 2016}\/} (pp.
  \bibinfo{pages}{118--127}).
\newblock \bibinfo{address}{Austin, Texas, USA}.
\bibitem[{Liu et~al.(2017)Liu, Qiu \& Huang}]{Liu17}
\bibinfo{author}{Liu, P.}, \bibinfo{author}{Qiu, X.}, \&
  \bibinfo{author}{Huang, X.} (\bibinfo{year}{2017}).
\newblock \bibinfo{title}{Adversarial multi-task learning for text
  classification}.
\newblock In {\it \bibinfo{booktitle}{Proceedings of the 55th Annual Meeting of
  the Association for Computational Linguistics, ACL 2017}\/} (pp.
  \bibinfo{pages}{1--10}).
\newblock \bibinfo{address}{Vancouver, Canada}.
\bibitem[{Liu \& Lapata(2018)}]{Liu18}
\bibinfo{author}{Liu, Y.}, \& \bibinfo{author}{Lapata, M.}
  (\bibinfo{year}{2018}).
\newblock \bibinfo{title}{Learning structured text representations}.
\newblock {\it \bibinfo{journal}{Transactions of the Association for
  Computational Linguistics}\/},  {\it \bibinfo{volume}{6}\/},
  \bibinfo{pages}{63--75}.
\bibitem[{Luong et~al.(2015)Luong, Pham \& Manning}]{luong2015effective}
\bibinfo{author}{Luong, T.}, \bibinfo{author}{Pham, H.}, \&
  \bibinfo{author}{Manning, C.~D.} (\bibinfo{year}{2015}).
\newblock \bibinfo{title}{Effective approaches to attention-based neural
  machine translation}.
\newblock In {\it \bibinfo{booktitle}{Proceedings of the 2015 Conference on
  Empirical Methods in Natural Language Processing, EMNLP 2015}\/} (pp.
  \bibinfo{pages}{1412--1421}).
\newblock \bibinfo{address}{Lisbon, Portugal}.
\bibitem[{Ma et~al.(2017)Ma, Li, Zhang \& Wang}]{ma2017interactive}
\bibinfo{author}{Ma, D.}, \bibinfo{author}{Li, S.}, \bibinfo{author}{Zhang,
  X.}, \& \bibinfo{author}{Wang, H.} (\bibinfo{year}{2017}).
\newblock \bibinfo{title}{Interactive attention networks for aspect-level
  sentiment classification}.
\newblock In {\it \bibinfo{booktitle}{Proceedings of the 26th International
  Joint Conference on Artificial Intelligence, IJACI 2017}\/} (pp.
  \bibinfo{pages}{4068--4074}).
\newblock \bibinfo{address}{Melbourne, Australia}.
\bibitem[{Van~der Maaten \& Hinton(2008)}]{VanDerMaaten08}
\bibinfo{author}{Van~der Maaten, L.}, \& \bibinfo{author}{Hinton, G.}
  (\bibinfo{year}{2008}).
\newblock \bibinfo{title}{Visualizing data using {t-SNE}}.
\newblock {\it \bibinfo{journal}{Journal of Machine Learning Research}\/},
  {\it \bibinfo{volume}{9}\/}, \bibinfo{pages}{2579--2605}.
\bibitem[{McAuley et~al.(2015)McAuley, Targett, Shi \& van~den
  Hengel}]{McAuley15}
\bibinfo{author}{McAuley, J.}, \bibinfo{author}{Targett, C.},
  \bibinfo{author}{Shi, Q.}, \& \bibinfo{author}{van~den Hengel, A.}
  (\bibinfo{year}{2015}).
\newblock \bibinfo{title}{Image-based recommendations on styles and
  substitutes}.
\newblock In {\it \bibinfo{booktitle}{Proceedings of the 38th International ACM
  SIGIR Conference on Research and Development in Information Retrieval, SIGIR
  2015}\/}.
\newblock \bibinfo{address}{Santiago, Chile}.
\bibitem[{Mcauliffe \& Blei(2008)}]{Blei08}
\bibinfo{author}{Mcauliffe, J.~D.}, \& \bibinfo{author}{Blei, D.~M.}
  (\bibinfo{year}{2008}).
\newblock \bibinfo{title}{Supervised topic models}.
\newblock In {\it \bibinfo{booktitle}{Advances in Neural Information Processing
  Systems 20, NIPS 2008}\/} (pp. \bibinfo{pages}{121--128}).
\newblock \bibinfo{address}{Vancouver, Canada}.
\bibitem[{Pennington et~al.(2014)Pennington, Socher \& Manning}]{Pennington14}
\bibinfo{author}{Pennington, J.}, \bibinfo{author}{Socher, R.}, \&
  \bibinfo{author}{Manning, C.} (\bibinfo{year}{2014}).
\newblock \bibinfo{title}{Glove: Global vectors for word representation}.
\newblock In {\it \bibinfo{booktitle}{Proceedings of the 2014 Conference on
  Empirical Methods in Natural Language Processing EMNLP 2014}\/} (pp.
  \bibinfo{pages}{1532--1543}).
\newblock \bibinfo{address}{Doha, Qatar}.
\bibitem[{Peters et~al.(2018)Peters, Neumann, Iyyer, Gardner, Clark, Lee \&
  Zettlemoyer}]{Peters18}
\bibinfo{author}{Peters, M.}, \bibinfo{author}{Neumann, M.},
  \bibinfo{author}{Iyyer, M.}, \bibinfo{author}{Gardner, M.},
  \bibinfo{author}{Clark, C.}, \bibinfo{author}{Lee, K.}, \&
  \bibinfo{author}{Zettlemoyer, L.} (\bibinfo{year}{2018}).
\newblock \bibinfo{title}{Deep contextualized word representations}.
\newblock In {\it \bibinfo{booktitle}{Proceedings of the 2018 Conference of the
  North American Chapter of the Association for Computational Linguistics:
  Human Language Technologies, NAACL 2018}\/} (pp.
  \bibinfo{pages}{2227--2237}).
\newblock \bibinfo{address}{New Orleans, Louisiana}.
\bibitem[{R\"{o}der et~al.(2015)R\"{o}der, Both \& Hinneburg}]{Roder15}
\bibinfo{author}{R\"{o}der, M.}, \bibinfo{author}{Both, A.}, \&
  \bibinfo{author}{Hinneburg, A.} (\bibinfo{year}{2015}).
\newblock \bibinfo{title}{Exploring the space of topic coherence measures}.
\newblock In {\it \bibinfo{booktitle}{Proceedings of the Eighth ACM
  International Conference on Web Search and Data Mining, WSDM 2015}\/} (pp.
  \bibinfo{pages}{399--408}).
\newblock \bibinfo{address}{Shanghai, China}.
\bibitem[{Saxe et~al.(2014)Saxe, McClelland \& Ganguli}]{SaxeMG13}
\bibinfo{author}{Saxe, A.~M.}, \bibinfo{author}{McClelland, J.~L.}, \&
  \bibinfo{author}{Ganguli, S.} (\bibinfo{year}{2014}).
\newblock \bibinfo{title}{Exact solutions to the nonlinear dynamics of learning
  in deep linear neural networks}.
\newblock In {\it \bibinfo{booktitle}{Proceedings of the 2015 International
  Conference for Learning Representations, ICLR 2014}\/}.
\newblock \bibinfo{address}{Banff, Canada}.
\bibitem[{Stab et~al.(2018)Stab, Miller, Schiller, Rai \& Gurevych}]{Stab18}
\bibinfo{author}{Stab, C.}, \bibinfo{author}{Miller, T.},
  \bibinfo{author}{Schiller, B.}, \bibinfo{author}{Rai, P.}, \&
  \bibinfo{author}{Gurevych, I.} (\bibinfo{year}{2018}).
\newblock \bibinfo{title}{Cross-topic argument mining from heterogeneous
  sources}.
\newblock In {\it \bibinfo{booktitle}{Proceedings of the 2018 Conference on
  Empirical Methods in Natural Language Processing EMNLP 2018}\/} (pp.
  \bibinfo{pages}{3664--3674}).
\newblock \bibinfo{address}{Brussels, Belgium}.
\bibitem[{Tang et~al.(2015)Tang, Qin \& Liu}]{Tang15}
\bibinfo{author}{Tang, D.}, \bibinfo{author}{Qin, B.}, \& \bibinfo{author}{Liu,
  T.} (\bibinfo{year}{2015}).
\newblock \bibinfo{title}{Document modeling with gated recurrent neural network
  for sentiment classification}.
\newblock In {\it \bibinfo{booktitle}{Proceedings of the 2015 Conference on
  Empirical Methods in Natural Language Processing EMNLP 2015}\/} (pp.
  \bibinfo{pages}{1422--1432}).
\newblock \bibinfo{address}{Lisbon, Portugal}.
\bibitem[{Van Der~Maaten(2014)}]{VanDerMaaten14}
\bibinfo{author}{Van Der~Maaten, L.} (\bibinfo{year}{2014}).
\newblock \bibinfo{title}{Accelerating t-{SNE} using tree-based algorithms}.
\newblock {\it \bibinfo{journal}{Journal of Machine Learning Research}\/},
  {\it \bibinfo{volume}{15}\/}, \bibinfo{pages}{3221--3245}.
\bibitem[{Vaswani et~al.(2017)Vaswani, Shazeer, Parmar, Uszkoreit, Jones,
  Gomez, Kaiser \& Polosukhin}]{vaswani2017attention}
\bibinfo{author}{Vaswani, A.}, \bibinfo{author}{Shazeer, N.},
  \bibinfo{author}{Parmar, N.}, \bibinfo{author}{Uszkoreit, J.},
  \bibinfo{author}{Jones, L.}, \bibinfo{author}{Gomez, A.~N.},
  \bibinfo{author}{Kaiser, {\L}.}, \& \bibinfo{author}{Polosukhin, I.}
  (\bibinfo{year}{2017}).
\newblock \bibinfo{title}{Attention is all you need}.
\newblock In {\it \bibinfo{booktitle}{Advances in Neural Information Processing
  Systems NIPS 2017}\/} (pp. \bibinfo{pages}{5998--6008}).
\newblock \bibinfo{address}{Long Beach, California, USA}.
\bibitem[{Wang et~al.(2018)Wang, Feng, Gao, Wang \& Zhang}]{Wang18}
\bibinfo{author}{Wang, W.}, \bibinfo{author}{Feng, S.}, \bibinfo{author}{Gao,
  W.}, \bibinfo{author}{Wang, D.}, \& \bibinfo{author}{Zhang, Y.}
  (\bibinfo{year}{2018}).
\newblock \bibinfo{title}{Personalized microblog sentiment classification via
  adversarial cross-lingual multi-task learning}.
\newblock In {\it \bibinfo{booktitle}{Proceedings of the 2018 Conference on
  Empirical Methods in Natural Language Processing EMNLP 2018}\/} (pp.
  \bibinfo{pages}{338--348}).
\newblock \bibinfo{address}{Brussels, Belgium}.
\bibitem[{Wang et~al.(2017)Wang, Yang, Wei, Chang \& Zhou}]{wang2017gated}
\bibinfo{author}{Wang, W.}, \bibinfo{author}{Yang, N.}, \bibinfo{author}{Wei,
  F.}, \bibinfo{author}{Chang, B.}, \& \bibinfo{author}{Zhou, M.}
  (\bibinfo{year}{2017}).
\newblock \bibinfo{title}{Gated self-matching networks for reading
  comprehension and question answering}.
\newblock In {\it \bibinfo{booktitle}{Proceedings of the 55th Annual Meeting of
  the Association for Computational Linguistics ACL 2017}\/} (pp.
  \bibinfo{pages}{189--198}).
\newblock \bibinfo{address}{Vancouver, Canada}.
\bibitem[{Wu \& Huang(2016)}]{Wu16}
\bibinfo{author}{Wu, F.}, \& \bibinfo{author}{Huang, Y.}
  (\bibinfo{year}{2016}).
\newblock \bibinfo{title}{Personalized microblog sentiment classification via
  multi-task learning}.
\newblock In {\it \bibinfo{booktitle}{Proceedings of the Thirtieth AAAI
  Conference on Artificial Intelligence AAAI 2016}\/} (pp.
  \bibinfo{pages}{3059--3065}).
\newblock \bibinfo{address}{Phoenix, Arizona}.
\bibitem[{Xu et~al.(2015)Xu, Ba, Kiros, Cho, Courville, Salakhudinov, Zemel \&
  Bengio}]{xu2015show}
\bibinfo{author}{Xu, K.}, \bibinfo{author}{Ba, J.}, \bibinfo{author}{Kiros,
  R.}, \bibinfo{author}{Cho, K.}, \bibinfo{author}{Courville, A.},
  \bibinfo{author}{Salakhudinov, R.}, \bibinfo{author}{Zemel, R.}, \&
  \bibinfo{author}{Bengio, Y.} (\bibinfo{year}{2015}).
\newblock \bibinfo{title}{Show, attend and tell: Neural image caption
  generation with visual attention}.
\newblock In {\it \bibinfo{booktitle}{International conference on machine
  learning ICML 2015}\/} (pp. \bibinfo{pages}{2048--2057}).
\newblock \bibinfo{address}{Lille, France}.
\bibitem[{Yang et~al.(2019)Yang, Zhang, Jiang \& Li}]{Yang2019}
\bibinfo{author}{Yang, C.}, \bibinfo{author}{Zhang, H.},
  \bibinfo{author}{Jiang, B.}, \& \bibinfo{author}{Li, K.}
  (\bibinfo{year}{2019}).
\newblock \bibinfo{title}{Aspect-based sentiment analysis with alternating
  coattention networks}.
\newblock {\it \bibinfo{journal}{Information Processing \& Management}\/},
  {\it \bibinfo{volume}{56}\/}, \bibinfo{pages}{463 -- 478}.
\bibitem[{Yang et~al.(2016)Yang, Yang, Dyer, He, Smola \& Hovy}]{Yang16}
\bibinfo{author}{Yang, Z.}, \bibinfo{author}{Yang, D.}, \bibinfo{author}{Dyer,
  C.}, \bibinfo{author}{He, X.}, \bibinfo{author}{Smola, A.}, \&
  \bibinfo{author}{Hovy, E.} (\bibinfo{year}{2016}).
\newblock \bibinfo{title}{Hierarchical attention networks for document
  classification}.
\newblock In {\it \bibinfo{booktitle}{Proceedings of the 2016 Conference of the
  North {A}merican Chapter of the Association for Computational Linguistics:
  Human Language Technologies NAACL 2016}\/} (pp. \bibinfo{pages}{1480--1489}).
\newblock \bibinfo{address}{San Diego, California}.
\bibitem[{Zhang et~al.(2018)Zhang, Xiao, Chen, Wang \& Jin}]{zhang18}
\bibinfo{author}{Zhang, H.}, \bibinfo{author}{Xiao, L.}, \bibinfo{author}{Chen,
  W.}, \bibinfo{author}{Wang, Y.}, \& \bibinfo{author}{Jin, Y.}
  (\bibinfo{year}{2018}).
\newblock \bibinfo{title}{Multi-task label embedding for text classification}.
\newblock In {\it \bibinfo{booktitle}{Proceedings of the 2018 Conference on
  Empirical Methods in Natural Language Processing EMNLP 2018}\/} (pp.
  \bibinfo{pages}{4545--4553}).
\newblock \bibinfo{address}{Brussels, Belgium}.
\bibitem[{Zheng et~al.(2018)Zheng, Chen \& Qiu}]{Zheng18}
\bibinfo{author}{Zheng, R.}, \bibinfo{author}{Chen, J.}, \&
  \bibinfo{author}{Qiu, X.} (\bibinfo{year}{2018}).
\newblock \bibinfo{title}{Same representation, different attentions: Shareable
  sentence representation learning from multiple tasks}.
\newblock In {\it \bibinfo{booktitle}{Proceedings of the Twenty-Seventh
  International Joint Conference on Artificial Intelligence, {IJCAI} 2018}\/}
  (pp. \bibinfo{pages}{4616--4622}).
\newblock \bibinfo{address}{Stockholm, Sweden}.

\end{thebibliography}

\end{document}